# Academic Resource Text Level Multi-label Classification based on Attention

Yue Wang[1], Yawen Li[2]*, Ang Li[1]

（1 School of Computer Science, Beijing Key Laboratory of Intelligent Telecommunication Software and Multimedia, Beijing University of Posts and Telecommunications, Beijing 100876, China）

（2 School of economics and management, Beijing University of Posts and Telecommunications, Beijing 100876, China）

**Abstract** Hierarchical multi-label academic text classification (HMTC) is to assign academic texts into a hierarchically structured labeling system. We propose an attention-based hierarchical multi-label classification algorithm of academic texts (AHMCA) by integrating features such as text, keywords, and hierarchical structure, the academic documents are classified into the most relevant categories. We utilize word2vec and BiLSTM to obtain embedding and latent vector representations of text, keywords, and hierarchies. We use hierarchical attention mechanism to capture the associations between keywords, label hierarchies, and text word vectors to generate hierarchical-specific document embedding vectors to replace the original text embeddings in HMCN-F. The experimental results on the academic text dataset demonstrate the effectiveness of the AHMCA algorithm.

**Keywords**    hierarchical multi-label classification; attention; BiLSTM; Word2Vec

With the development of scientific and technological big data, more and more scientific and technological resources (papers and patents) appear in the network [1]. Due to the wide range of subject areas involved in scientific and technological resources, the classification system often has a hierarchical structure, and the classification of scientific and technological resources texts is suitable for the use of hierarchical multi-label classification. At the same time, compared with ordinary text classification, scientific and technological resource texts often have keyword fields [2]. How to make good use of keyword fields to optimize the classification of scientific and technological resource texts is a problem that needs to be solved. Hierarchical multi-label classification has many application scenarios in the real world, such as document classification [3], news classification [4], text classification of scientific and technological resources [5] etc. Its category information (label) is a system with a hierarchical relationship, and automatic classification of them involves here. Selecting multiple correct labels in the hierarchical labeling system constitutes the problem of hierarchical multi-label text classification. How to learn and utilize these different hierarchical relationships and evaluate the classification results from the perspective of hierarchical relationship compliance has become the difficulty and challenge of hierarchical multi-label classification.

This paper conducts related research on the text-level multi-label classification of scientific and technological resources. Based on the HMCN-F [6] and the keywords of scientific and technological resources, an attention-based text-level multi-label text classification method for scientific and technological resources (AHMCA) is proposed. Utilize word2vec and BiLSTM to obtain embedding vector and latent vector representation of text, keywords, and hierarchy. Then, based on the attention mechanism, the similarity weight between keywords, label hierarchy and text word vectors is captured, and the layer-specific document embedding vector is obtained based on the similarity weight generated under different level labels, which replaces the original text embedding in HMCN-F.

The main contributions of this paper include:

1) A hierarchical multi-label classification algorithm for scientific and technological resources text based on attention mechanism is proposed. The hierarchical attention mechanism is used to capture the relationship between text word vectors, keywords, and label hierarchies to generate hierarchical-specific documents. Embedding vector.

2) The correlation matrix between the keywords of

This work is supported by National Key R&D Program of China (2018YFB1402600), the National Natural Science Foundation of China (61772083, 61802028, 61877006, 62002027).

**Corresponding Author**：Yawen Li（warmly0716@126.com）

scientific and technological resources and the word vectors of the text is established, and the weight of the word vectors related to the keywords in the text of scientific and technological resources is strengthened. The context information is fused by BiLSTM, which strengthens the method's attention to key word vectors while retaining the context information.

3) For inputs of different levels, we construct a level-specific document embedding vector according to the level label vector of each layer, and calculate the weight of each word vector by constructing the similarity matrix between the level label vector and the text word vector, and by weighting the hierarchical document embedding vectors are calculated on average.

## 1 Related Works

There has been a lot of work on the problem of hierarchical multi-label classification [7][8][9]. Initially, it was proposed to use multi-label classification methods such as Naive Bayes [10] for hierarchical classification, by simplifying the hierarchical multi-label classification problem to a flattened multi-label problem, only predicting the last-level category. However, these simple methods ignore hierarchical category structure information and perform poorly [11][12]. Subsequently, some methods consider the hierarchical information of labels, which are divided into local and global methods according to the adopted strategy. Regarding local methods, Cesa-Bianchi [13] proposed a classification method using a hierarchical SVM, where the SVM only starts learning when its parent class is marked as positive. Rousu et al. proposed a margin method [14][15] to compute the maximum structural margin of the output class. On the global approach, Vens et al. [16] developed a tree-based approach called Clus-HMC to handle the entire hierarchical category structure. Cerri et al. [17] attempted to use HMC-LMLP to incrementally train a set of neural networks [18][19], each responsible for predicting a class at a given level. Borges [20] proposed a global approach based on adversarial neural networks [21] to predict all categories in a hierarchy. Schietgat [22] proposed an inheritance-improved HMC based on Clus-HMC and introduced decision tree. Kiritchenko [23] proposes a global method based on AdaBoost.MH [24]. Kim [25] uses Word2vec as the model input, and then uses windows of different sizes as feature extractors to perform convolution operations to obtain text representations at different levels. Peng [26] used Graph-of-Words to capture non-continuous and long-distance semantics in text.

However, the above works mainly focus on the overall structure of local regions or class labels. Dependencies between different levels of the hierarchy are ignored, which leads to downward propagation of mispredictions and class membership inconsistencies [27][28]. Hybrid methods combining the advantages of local and global methods have been applied in many domains, among them, HMCN [6][29] integrates the prediction results of each level in the hierarchy and the overall hierarchy. However, HMCN fails to capture the association between text and hierarchy.

One of the most important steps of the hierarchical multi-label classification method is to reveal the association between the text and each category in the hierarchy from top to bottom, and to give sufficient weight to the text word vector, which requires the use of attention mechanism [30][31]. Huang [32] used an attention mechanism on top of the CNN model to introduce an additional source of information to guide the extraction of sentence embeddings. Tao [33] utilizes an attention method that creates a weighted vector representation by using the encoding of the hidden states of any RNN. Lin [34] designed a self-attention mechanism for sequence models such as RNNs instead of max-pooling or averaging steps, which enables the model to focus on different aspects of sentences.

## 2 Definition of text-level multi-label classification of scientific and technological resources

In the text-level multi-label classification of scientific and technological resources, for a set of documents, each document contains title, abstract and keywords, and the expected classification result of the document is organized into a hierarchical structure. The relevant definitions of the hierarchical structure are given below.

**Definition 1** Hierarchical structure $\gamma$. Define the category set $\mathbb{C} = (C^1, C^2, ..., C^H)$, where $H$ is the depth of the hierarchy. $C^i = \{c_1, c_2, ...\} \in \{0,1\}^{|C^i|}$ is the possible classes on the i-th layer, $|C^i|$ is the number of classes in the current layer, and $K$ is the total number of classes. Define a hierarchy $\gamma$ on $\mathbb{C}$ as a partially ordered set $(\mathbb{C}, \prec)$. $\prec$ stands for PARENT-OF partial order, which is asymmetric, reflexive and transitive:

$$\forall c_x \in C^i, c_y \in C^j, c_x \prec c_y \text{ then } c_y \not\prec c_x$$
$$\forall c_x \in C^l, c_x \not\prec c_x$$
$$\forall c_x \in C^i, c_y \in C^j, c_z \in C^k, if\ c_x \prec c_y\ and\ c_y \prec c_z\ then\ c_x \prec c_z$$

A set of M documents accompanied by labels can be represented as $X = \{(D_1, L_1), (D_2, L_2) ... (D_M, L_M)\}$, where $D_i = \{w_1, w_2, ..., w_N\}$ denoted by N A text sequence composed of words, $L_i = \{\ell_1, \ell_2, ..., \ell_H\}$ represents the label set of $D_i$, $\ell_i \subset C^i$. The HMTC problem can be formulated.

**Definition 2** Hierarchical multi-label classification of scientific and technological resources. Given a set of documents and the corresponding hierarchical label structure, the goal is to use D and the corresponding label structure $\gamma$ to learn a classification model $\Omega$ that can predict the hierarchical classification L:

$$\Omega(D, \gamma, \Theta) \to L \quad (1)$$

where $\Theta$ is the parameter of $\Omega$.

## 3 Hierarchical multi-label classification algorithm of scientific and technological resources text based on attention mechanism

The main framework of the AHMCA algorithm, as shown in Figure 1, AHMCA mainly consists of two parts, the first part is the document representation and attention mechanism layer, and the second layer is the HMCN-F layer. The document representation and attention mechanism layer is used to capture the relationship between the text corpus and keywords, hierarchical labels, and contextual information of the text. The generated hierarchical document embedding vector is fed into HMCN-F for hierarchical label prediction. The main implementation steps of the algorithm are shown in Table 1.

Table 1 Algorithm steps of multi-label classification of scientific and technological resources text based on attention mechanism

| Hierarchical multi-label classification algorithm of scientific and technological resources text based on attention mechanism |
|---|
| Input: text of scientific and technological resources (title, abstract), keywords |
| Output: Classification result |
| (1) Use Word2Vec to embed text (title and abstract), keywords, and level tags of scientific and technological resources |
| (2) The title, abstract, and keywords of the text are spliced together and sent to BiLSTM, and the hidden layer vectors $\vec{H}$ and $\overleftarrow{H}$ are extracted. |
| (3) Splicing the keyword vector matrix under each level label matrix, constructs the level label splicing matrix $T_c^i$, $T_c^i = [T^i, K_e]$. |
| (4) For calculating the similarity between the hidden layer vector and the $T_c^i$ vector under each level, construct the hidden layer vector weight matrix. The specific formula is as follows: $$\vec{a_j^i} = max_l(\vec{h_j} \cdot t_l)$$ $$\vec{a^i} = [\vec{a_1^i}, ..., \vec{a_n^i}]$$ $$\overleftarrow{a_j^i} = max_l(\overleftarrow{h_j} \cdot t_l)$$ $$\overleftarrow{a^i} = [\overleftarrow{a_1^i}, ..., \overleftarrow{a_n^i}]$$ |
| (5) Based on the hidden layer vector weights in both directions, construct a hierarchically specialized text embedding vector $x^i$: $$\vec{x^i} = \vec{a^i} \cdot \vec{H}$$ $$\overleftarrow{x^i} = \overleftarrow{a^i} \cdot \overleftarrow{H}$$ $$x^i = [\vec{x^i}; \overleftarrow{x^i}]$$ |
| (6) Input the hierarchically specialized text embedding vector into each level input of the HMCN-F model for hierarchical multi-label classification |

**3.1** Text representation and attention mechanism layer of scientific and technological resources

In the first stage of the AHMCA method, the document representation layer generates a unified vector representation of text, keywords, and hierarchical labels. To achieve this goal, Word2Vec is used to encode text (title and abstract), keywords, and hierarchical labels. Then, the title, abstract, and keywords of the text are spliced together and sent to BiLSTM to strengthen the semantic feature representation of the text. In the attention mechanism layer, the keyword vector matrix is spliced under each level label matrix, and the level label splicing matrix $T_c^i$ (representing the splicing matrix of the i-th level label matrix and the keyword vector matrix) is constructed. By calculating $T_c^i$ The similarity between each vector in i and the set of latent vectors ($\vec{H}$ and $\overleftarrow{H}$ ) output by BiLSTM obtains the word weights in both directions ($\overrightarrow{a^i}$ and $\overleftarrow{a^i}$), respectively The word weights are used to weighted the latent vectors in the corresponding directions to obtain the document embedding representations in the two directions, and the document embedding representations in the two directions are spliced together to obtain the level-specific document embedding representation $x_i$.

The word embedding layer encodes individual words in the text, hierarchical labels, and keywords. Taking document D containing abstracts and titles, hierarchical labels $\gamma$, and keyword set K as input (as shown in Figure 1), documents containing N words $D = (w_1, w_2, ..., w_N)$, containing M words Keywords $K = (w_1, w_2, ..., w_M)$, where $w_i \in \mathbb{R}^k$ is the word2vec pre-training model encoded as a k-dimensional word vector. Use word2vec to embed the hierarchical label $\gamma$ as a hierarchical label matrix set $T = (T^1, T^2, ..., T^H)$, where $T^i \in \mathbb{R}^{|C^i| \times k}$ represents the i-th layer Hierarchical labels are the set of $|C^i|$ word vectors obtained by word2vec. If a label text is composed of multiple words, the mean vector of these word vectors is taken as the label vector.

The BiLSTM layer strengthens the semantic representation of text. For a document set D constructed from titles, abstracts, and keywords, BiLSTM is used to generate latent vector representations ($\overrightarrow{h_n}$ and $\overleftarrow{h_n}$) of each word in D. BiLSTM is an improvement of traditional LSTM[15,17]. BiLSTM can not only learn long-range semantic dependencies, but also aggregate forward and backward contextual information at the same time. $D = (w_1, w_2, ..., w_N)$, the calculation process of the hidden vector is as follows:

$$\overrightarrow{h_n} = \text{LSTM}(\overrightarrow{h_{n-1}}, w_n)$$
$$\overleftarrow{h_n} = \text{LSTM}(\overleftarrow{h_{n+1}}, w_n) \quad (2)$$

Where $\overrightarrow{h_n}$ and $\overleftarrow{h_n}$ are the forward and backward hidden vectors of the nth word vector $\overleftarrow{h_n}, \overrightarrow{h_n} \in \mathbb{R}^k$, as shown in Figure 1, $\overrightarrow{H} = (\overrightarrow{h_1}, \overrightarrow{h_2}, ..., \overrightarrow{h_n}), \overleftarrow{H} = (\overleftarrow{h_1}, \overleftarrow{h_2}, ..., \overleftarrow{h_n})$

The attention mechanism layer of the technology resource layer constructs the label splicing matrix of each level, and $T_c^i$ represents the label splicing matrix of the ith level:

$$T_c^i = [T^i, K_e] \quad (3)$$

Where $K_e$ represents the keyword vector matrix after word2vec embedding. Let $T_c^i = (t_1, t_2, ..., t_q)$, then the calculation method of the weight vector of the forward and backward hidden vectors of the i-th layer is as follows:

$$\overrightarrow{a_j^i} = max_l(\overrightarrow{h_j} \cdot t_l) \quad (4)$$

$$\overrightarrow{a^i} = [\overrightarrow{a_1^i}, ..., \overrightarrow{a_n^i}] \quad (5)$$

$$\overleftarrow{a_j^i} = max_l(\overleftarrow{h_j} \cdot t_l) \quad (6)$$

$$\overleftarrow{a^i} = [\overleftarrow{a_1^i}, ..., \overleftarrow{a_n^i}] \quad (7)$$

For each hidden layer word vector, the maximum similarity between the vector and each vector in $T_c^i$ is taken as the vector weight, and the hidden layer vector in each direction is calculated once. By using the forward and reverse weight vectors, you can The hidden vectors output by BiLSTM are weighted and averaged to obtain hierarchically specialized text embedding vectors:

$$\overrightarrow{x^i} = \overrightarrow{a^i} \cdot \overrightarrow{H} \quad (8)$$
$$\overleftarrow{x^i} = \overleftarrow{a^i} \cdot \overleftarrow{H} \quad (9)$$
$$x^i = [\overrightarrow{x^i}; \overleftarrow{x^i}] \quad (10)$$

As shown in Figure 1, $x^0$ is used for global label prediction without any weight attached to it, so $\overleftarrow{a^0} = \overrightarrow{a^0} = [1, ..., 1]$

### 3.2 Hierarchical multi-label classification of scientific and technological resources

As shown in Figure 1, the multi-label classification layer of the scientific and technological resource hierarchy contains two information flows. The global information flow corresponds to all information transmission, which is transmitted from the input $x^i$ to each intermediate layer $A_G^i$, and finally passes through a fully connected layer. Get the global hidden layer representation $P_G$. The local information flow corresponds to the information transmission flow of each intermediate layer, and the hidden layer representation $P_L^i$ of each layer is obtained through the local classification layer. The local hidden layer representation vector is spliced and then summed with the global hidden layer representation vector as the final Prediction. The global information flow can transfer semantic information from the i-th layer to the i+1-th layer, and the local information flow will also play a role in the intermediate layer representation in the global information flow through back-propagation.

The input is $x^i \in \mathbb{R}^{2k}$, which represents the semantic representation vector extracted from the text. For the global information flow, the input $A_G^h$ of each intermediate layer is expressed as:

$$A_G^h = \phi(W_G^h(A_G^{h-1} \odot x^h) + b_G^i) \quad (11)$$

The input is the result of the splicing of the output $A_G^{h-1}$ and $x^h$ of the previous layer. When h=1, the input is only $x^1$. The final output $P$ is represented as:

$$P_G = \sigma(W_G^{|H|+1} A_G^H + b_G^{|H|+1}) \quad (12)$$

where $P_G \in R^{|C|}$, $\sigma$ is the activation function, and each dimension represents the probability that $P_G^{(i)}$ belongs to the ith category.

For local flow, the intermediate output of the global flow needs to be mapped to the hidden layer output of the local layer:

$$A_L^h = \phi(W_T^h A_G^h + b_T^h) \quad (13)$$

And map it to the corresponding layer through a fully connected layer to get $P_L^h$:

$$P_L^h = \sigma(W_L^h A_L^h + b_L^h) \quad (14)$$

Where $P_L^h \in R^{|C^h|}, W_L^h \in R^{|C^h|*|A_L^h|}$, the final prediction output:

$$P_F = \beta(P_L^1 \odot P_L^2 \odot ... P_L^{|H|}) + (1-\beta)P_G \quad (15)$$

Among them, β is an adjustment hyperparameter, which is used to adjust the importance of local information and global information. The default setting is 0.5, which means equal importance.

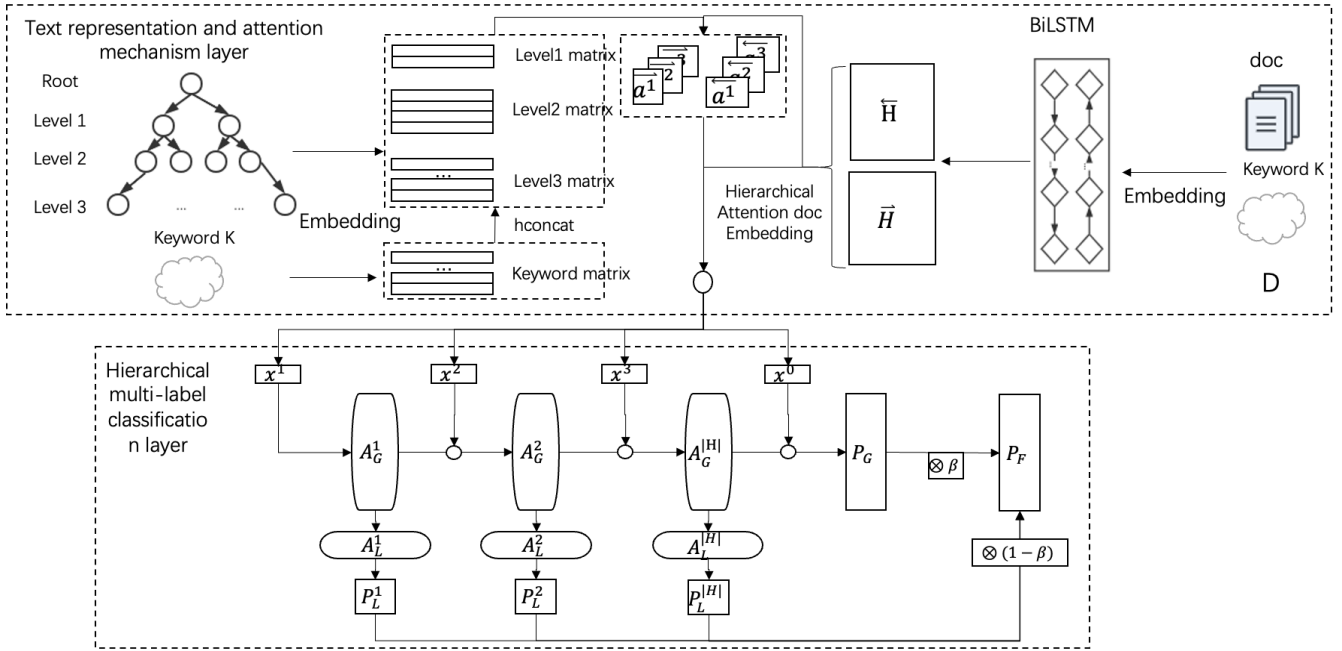

Figure 1 Framework of multi-label classification method of scientific and technological resources text based on attention mechanism

## 4 Experimental results
### 4.1 DataSet

This paper uses two datasets to verify the effectiveness of this method, the first dataset is a paper dataset and the second dataset is a patent dataset. The paper dataset contains title, abstract, keyword fields and the corresponding CLC hierarchical label structure. The patent dataset contains title and abstract, but does not contain keyword fields, corresponding to the International Patent Classification.

The paper dataset contains 70,829 papers. The CLC number is used as the hierarchical label, the level depth is 2, the number of labels for leaf nodes is 248, and the average number of papers for secondary labels is 531, which are divided according to the ratio of 3:1:1 for training set, validation set and validation set.

The patent dataset contains 23,123 patents, using the International Patent Classification Number as the hierarchical label, the hierarchical depth is 2 layers, the number of labels of leaf nodes is 95, and the average number of articles of secondary labels is 273, according to the ratio of 3:1:1 Divide into training set, validation set and validation set.

### 4.2 Evaluation Metrics

We uses the precision rate (P), recall rate (R), and F1-marco indicators to evaluate the final effect of the hierarchical classification of scientific and technological resources text:

$$P = \frac{1}{n}\sum_{1}^{n} P_i \quad (16)$$

$$R = \frac{1}{n}\sum_{1}^{n} R_i \quad (17)$$

$$\text{F1-marco} = \frac{2 \times P \times R}{P + R} \quad (18)$$

### 4.3 Contrast method

We use the following methods as a comparison to verify the performance of the AHMCA method:

TextRNN [35] uses RNN recurrent neural network to encode the original text vector, and then input it to the fully connected layer, and finally gives the classification label prediction result of the text.

AttConvNet [36] is a convolutional network with attention mechanism for extracting important information in text.

DPCNN [37] is a deep text classification convolutional neural network that captures long-range dependencies of text by deepening the network.

HMCN [1] integrates the prediction results of each level in the hierarchy and the overall hierarchy for hierarchical multi-label classification.

### 4.4 Experimental results and analysis

The comparison experiments of the above methods are carried out on the papers and patent data sets respectively, and the accuracy of each method in the top1, top3, and top5 cases and the F1 value in the top1 case are counted.

Table 2 Comparative experiment of AHMCA method on paper dataset

|  | P@1 | P@3 | P@5 | Marco-F1@1 |
|---|---|---|---|---|
| TextRNN | 0.8582 | 0.4379 | 0.2414 | 0.6266 |
| AttConvNet | 0.8274 | 0.4025 | 0.2383 | 0.5965 |
| DPCNN | 0.8045 | 0.3809 | 0.2681 | 0.5319 |
| HMCN | 0.8771 | 0.5135 | 0.2191 | 0.6894 |
| AHMCA | **0.8952** | **0.5614** | **0.3563** | **0.7052** |

As shown in Table 2, the results of the comparative experiments on the dataset of the paper are shown. It can be seen that the AHMCA method proposed in this paper is superior to the HMCN, DPCNN, AttConvNet, and TextRNN methods in four performance indicators. Compared with the HMCN method, the AHMCA method has about 2% improvement on Marco-F1. Among them, the HMCN method combines the local information and global information of the hierarchical labels, and its performance indicators are better than traditional flat multi-label classifiers such as TextRNN, AttConvNet, DPCNN, etc. The AHMCA method combines keywords and hierarchical labels on the basis of the HMCN method. The attention mechanism generates a hierarchically specialized document embedding vector, which integrates the hierarchical label weight and the keyword weight, and optimizes the input of the hierarchical label prediction, so it is better than the HMCN method in effect. TextRNN and others use the RNN structure as a semantic encoding layer to capture contextual information, while AttConvNet uses an attention mechanism based on CNN to capture non-nearest neighbor information, so that each word has higher-level features. Compared with DPCNN, these two methods have better performance on the paper dataset. As the top-n index decreases, the accuracy of the method decreases rapidly, mainly because most of the paper data has only one level label, and there are few cross-domain text instances, so it is difficult to obtain effective training data.

Table 3 Comparative experiment of AHMCA method on patent dataset

|  | P@1 | P@3 | P@5 | Marco-F1@1 |
|---|---|---|---|---|
| TextRNN | 0.8014 | 0.4501 | 0.2593 | 0.5988 |
| AttConvNet | 0.7625 | 0.4283 | 0.2451 | 0.5162 |
| DPCNN | 0.7853 | 0.4462 | 0.2514 | 0.5535 |
| HMCN | 0.8126 | 0.4963 | 0.3086 | 0.6062 |
| AHMCA | **0.8249** | **0.5315** | **0.3282** | **0.6391** |

Table 3 shows the comparative experimental results of patent data sets. Patent data and paper data have similarities in grammatical structure and corpus, but patent data lacks keyword fields, so only hierarchical label-based attention is used during the experiment. force mechanism to construct text embedding vectors. It can be seen that the AHMCA method outperforms other flat multi-label separation methods on the patent dataset. Compared with the HMCN method, the performance of Marco-F1 has been improved by 3%. The related methods that use the CNN coding layer such as AttConvNet and DPCNN have poor results, indicating that compared with CNN, RNN can be well extracted in text classification. The underlying semantic information in the text. The AHMCA method uses BiLSTM to capture the context-dependent information of scientific and technological resource texts, and uses different levels of label vectors to weight the key vectors in the text, optimizing the document vector output by the encoding layer. At the same time, the multi-label classification layer of the scientific and technological resource level comprehensively utilizes the overall information and local information of the hierarchical labels, and gives the hierarchical penalty loss function to reduce the probability of the hierarchical classifier's prediction error. Therefore, AHMCA is relatively flat RNN classifier in Marco- There is a 4% performance improvement on F1.

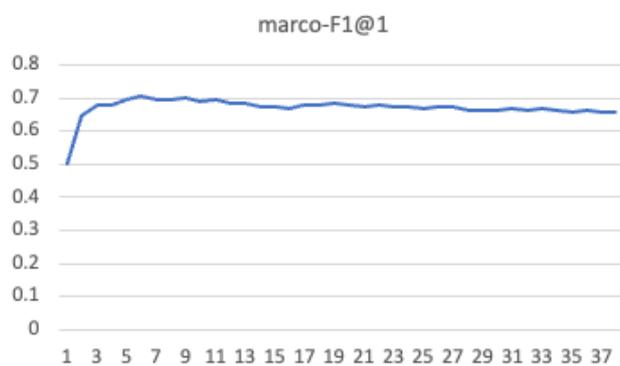

Figure 4 marco-F1 is affected by the number of iterations

Figure 4 shows the change of marco-f1 of the method on the test set with the increase of the number of iterations of the AHMCA method on the paper dataset. It can be seen that when the iteration range is between [1, 7], with the increase of the number of iterations As the number of iterations increases, the F1 value on the test set gradually increases from 0.5 to 0.7, indicating that the information fitted by AHMCA from the training set improves the overall effect of the method. However, as the number of iterations exceeds 10, the F1 value on the test set gradually decreases and stabilizes between 0.6 and 0.7, indicating that the AHMCA method gradually begins to overfit the training set, and the performance on the test data set has declined. Therefore, under the same dataset size, the number of iterations should be controlled within 20 to save time and computing resources.

## 5 Conclusion

This paper proposes a hierarchical multi-label classification algorithm for scientific and technological resources text based on attention mechanism. The hierarchical attention mechanism is used to capture the relationship between text word vectors, keywords and label hierarchies to generate hierarchical-specific document embeddings. vector. By establishing the correlation matrix between the keywords of scientific and technological resources and the text word vectors, the weight of the word vectors related to the keywords is strengthened, and BiLSTM is used to fuse the context information. While retaining context information, the method's attention to key word vectors is enhanced. For inputs of different levels, construct a level-specific document embedding vector according to the level label vector of each layer, and calculate the weight of each word vector by constructing the similarity matrix between the level label vector and the text word vector, and calculate by weighted average Out-level document embedding vector. The experimental results verify the effectiveness of the proposed method in the text-level multi-label classification of scientific and technological resources.

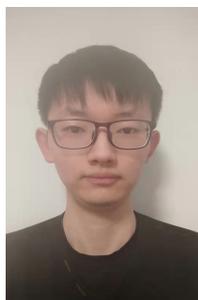

**Yue Wang** was born in 1997, is a Master candidate in Computer Science of Beijing University of Posts and Telecommunications. His research interests include nature language processing, data mining and deep learning.

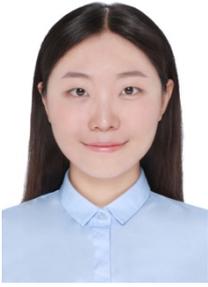
**Yawen Li** (corresponding author) was born in 1991. She is now an associate professor of School of economics and management, Beijing University of Posts and telecommunications. The main research directions are enterprise innovation, artificial intelligence, big data, etc.（warmly0716@126.com）

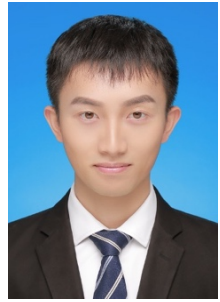
**Ang Li** was born in 1993. He is currently working toward the Ph.D. degree in Computer Science and Technology at the Beijing University of Posts and Telecommunications, China. His major research interests include information retrieval, data mining and machine learning.